\documentclass[letterpaper, 10 pt, conference]{ieeeconf}  

\IEEEoverridecommandlockouts                              

\usepackage{graphics} 
\usepackage{amsmath} 
\usepackage{todonotes}
\usepackage[bookmarks=true]{hyperref}
\hypersetup{%
  breaklinks,
  bookmarksnumbered=true,
  bookmarksopen=true,
  colorlinks=true,
  linkcolor=black,
  urlcolor=black,
  citecolor=black
}

\usepackage{cite}

\title{\LARGE \bf
The Simplest Walking Robot: A bipedal robot with one actuator and two rigid bodies
}

\author{James Kyle$^{1}$, Justin K. Yim$^{2}$, Kendall Hart$^{1}$, Sarah Bergbreiter$^{1}$, and Aaron M. Johnson$^{1}$
\thanks{*This work was supported in part by the National Science Foundation under Grant IIS-1813920 and Grant CCF-2030859 to the Computing Research Association for the CIFellows Project.}
\thanks{$^{1}$Department of Mechanical Engineering, Carnegie Mellon University, 
Pittsburgh, PA 15232, USA, {\tt\small amj1@cmu.edu}}%
\thanks{$^{2}$Department of Mechanical Science and Engineering, University of Illinois Urbana-Champaign, Urbana-Champaign, IL 61801, USA}%
}

\begin{document}
\bstctlcite{IEEEexample:BSTcontrol}

\maketitle
\thispagestyle{empty}
\pagestyle{empty}

\begin{abstract}

We present the design and experimental results of the first 1-DOF, hip-actuated bipedal robot.
While passive dynamic walking is simple by nature, many existing bipeds inspired by this form of walking are complex in control, mechanical design, or both.
Our design using only two rigid bodies connected by a single motor aims to enable exploration of walking at smaller sizes where more complex designs cannot be constructed.
The walker, ``Mugatu'', is self-contained and autonomous, open-loop stable over a range of input parameters, able to stop and start from standing, and able to control its heading left and right. We analyze the mechanical design and distill down a set of design rules that enable these behaviors.
Experimental evaluations measure speed, energy consumption, and steering.

\end{abstract}

\section{Introduction}

Small robots have the potential to conduct inspection and maintenance inside machinery spaces that other robots cannot reach, explore small crevices on other planetary bodies, or search through rubble in disaster scenarios \cite{ryanReview, koleoso2020micro}.
Small walking robots in particular offer unique capabilities like careful step placement or hopping to navigate obstacles, which can be particularly useful at small scales where surfaces are comparatively rough and obstacles are comparatively large.

Achieving capable terrestrial locomotion at these scales is still an open challenge in part due to the difficulty of building small, complex hardware.
Here, we study simpler designs that can enable construction of small walking robots.
Furthermore, simple locomotion mechanisms can provide insight into the fundamental science of locomotion and reduce the cost and complexity of robot hardware.
This leads us to our research question: \emph{What is the simplest way to achieve walking?}

\begin{figure}[t]
\centering
\includegraphics[width=\linewidth,clip]{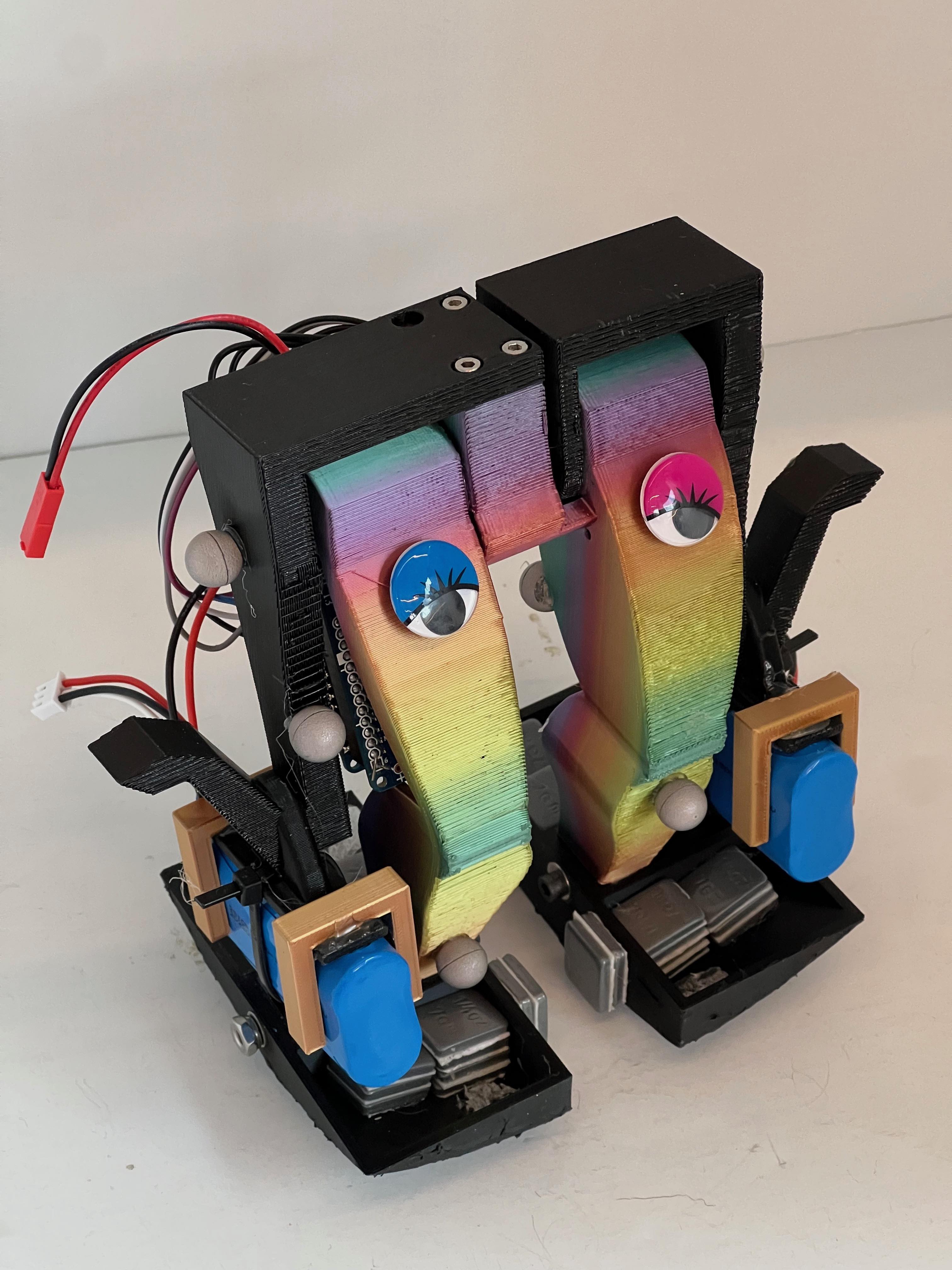}
\caption{The ``simplest" hip-actuated bipedal robot consisting of only two rigid bodies and containing only one actuator, exhibiting open-loop stable walking, self-starting, and left and right steering.}
\label{fig:robot}
\end{figure}

In this paper, we present the mechanical design, necessary parameter relationships, and experimental evaluation of what the authors believe may be one answer to this question. Specifically, we propose the first bipedal walking robot that contains only a single motor and is:
\begin{itemize}
    \item self-contained (power-autonomous)
    \item self-starting (from standing unsupported on flat ground)
    \item open-loop stable in its gait
    \item capable of controlled steering straight, left, and right
\end{itemize}
The robot prototype is shown in Fig.~\ref{fig:robot}, and 
in Table \ref{walkerCompare} we compare previous robots to this current robot with these metrics in mind.
The supplemental video with demonstrations of walking and turning can be found through the following YouTube link: \url{https://youtu.be/5EwBtk0PADw}.

\subsection{Related work}

Small robots are often simple by necessity due to restrictions in construction and components.
Most very small robots are not fully self-contained and are either externally actuated by external magnets or use offboard electrical power \cite{ryanReview}.
Among self-contained robots, some are not steerable \cite{bhushan2019insect} and others are steerable using two or more motors like their larger counterparts \cite{caprari2000fascination, sabelhaus2013}.
The small wheeled robot, Squirt, uses only one motor but can only drive straight forward or turn to one side while reversing \cite{flynn1989world}.
Other small robots use many legs or bristles, which can be actuated independently \cite{goldberg2018} to provide a statically stable stance, but increase the complexity and number of actuators. These multiple legs can also be synchronized \cite{microBristle, piezoHexapod, Pierre2016} at the cost of increasing the complexity of possible contact modes that cannot be easily controlled.

Passive dynamic walkers (PDWs) are a fascinating class of devices that require no actuation at all to produce stable gaits on sloped surfaces \cite{mcgeer, wilson1938walking}.
Many robots inspired by these mechanisms demonstrate impressive efficiency while using few actuators \cite{tedrake, collins2005efficient, wisse2007passive, ranger, kinugasa}.
However, most walkers in this area of research are either large in size, complex in control/actuation scheme, limited in forward direction control, or a combination of all three. See Table~\ref{tab:comparison} for a summary of some of these platforms.

Various theoretical models, from the ``simplest walking model'' \cite{garcia1998simplest} parameterized only by slope, or the closely related rimless wheel model \cite{mcgeer1989wobbling} to many models with curved feet \cite{tedrake, kuo1999}, describe motion inspired by PDWs.
Validating these models across different scale regimes would benefit from experiments with simple hardware designs that can be built at a range of scales.

Attempting to simplify these passive dynamics inspired robots, we previously designed ``Squeaky", a bipedal walking robot that uses five main rigid bodies with a passive hip joint and one telescopic joint per leg acting along the length of each leg \cite{islam2022}.
Despite the simplicity of this design, there is still room to decrease the number of actuators (two) and internal degrees of freedom (four) to only one, resulting in a robot composed of only two rigid bodies.
Here, we consider one method of actuating such a walker through a revolute hip joint at the intersection of two rigid legs.
We investigate a control scheme that is both challenging, since there are no actuators devoted to raising the feet, and attractive for its simplicity.



\begin{table}[t]
\vspace{.5em}
\caption{Comparison of different passive dynamics inspired walkers.}
\begin{center}
\label{walkerCompare}
\setlength{\tabcolsep}{0.5em}
\begin{tabular}{p{1.5cm}|c c c c c c}
 & Ranger & Collins & MIT &  RW04  & Squeaky & \textbf{Mugatu}\\
\hline
Citation & \cite{ranger} & \cite{collins2005bipedal} & \cite{tedrake} & \cite{kinugasa} & \cite{islam2022}  & \textbf{This work}\\
Mass (kg) & 9.9 & 12.7 & 2.9 & 6.5 & 0.365 & 0.809 \\
Length (m) & 1 & 0.81 & 0.44 & 0.807 & 0.15 & 0.15\\
Actuators & 3 & 2 & 4 & 2 & 2 & 1 \\
Open loop? & No & No & Yes & Yes & Yes & Yes\\
CoT & 0.19 & 0.20 & - & - & - & 5.3\\
Speed (cm/s) & 59 & 44 & - & - & 15 & 16\\
\end{tabular}
\label{tab:comparison}
\end{center}
\vspace{-0.2in}
\end{table}

\section{Methods}
\begin{figure}[t]
\vspace{.5em}
\centering
\includegraphics[width=0.95\linewidth]{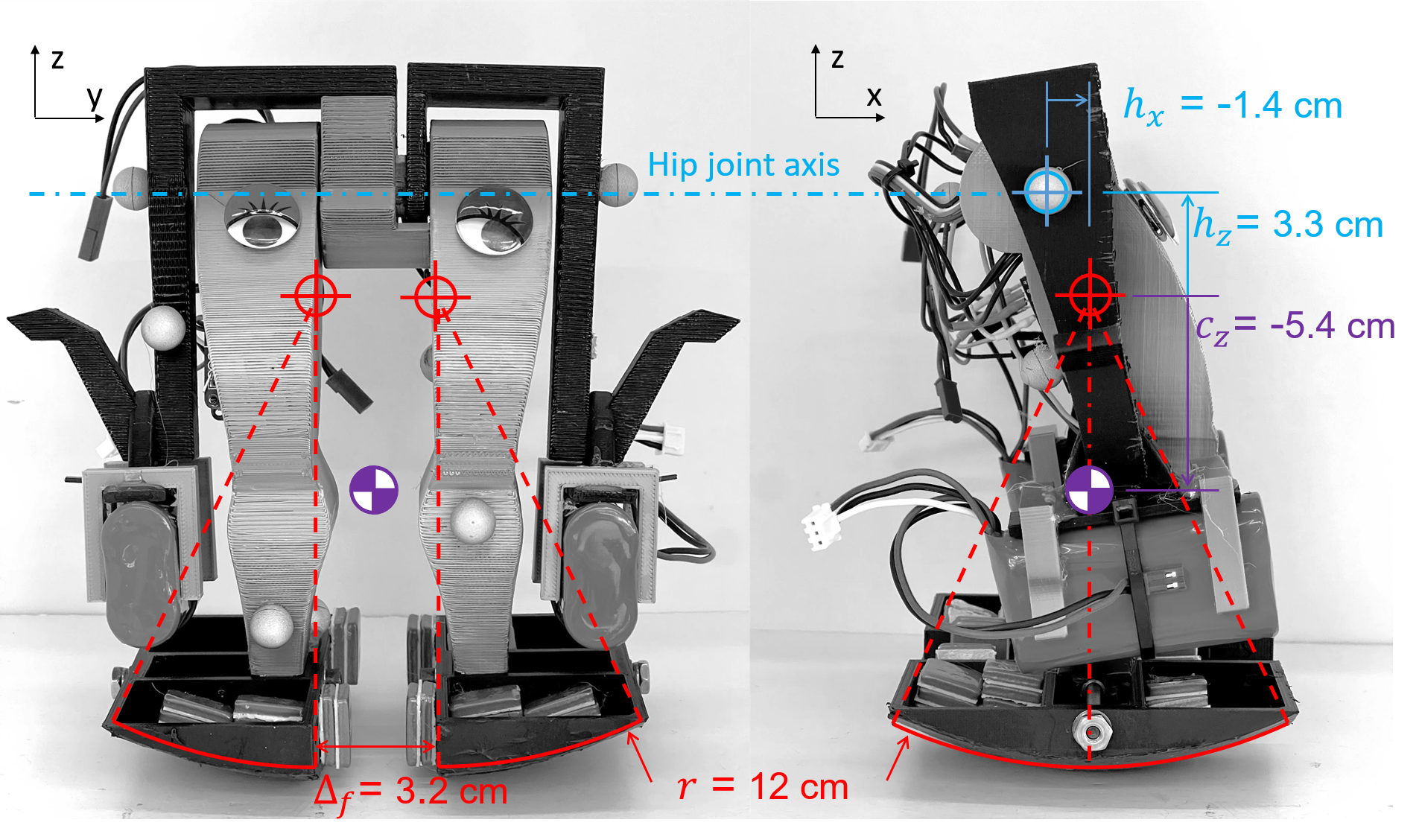}
\caption{Robot frontal and sagittal plane geometry.}
\label{fig:geometry}
\end{figure}

\begin{table}[t]
\caption{Design rules}
\centering
\begin{tabular}{l | c l}
No. & expression & description \\
\hline
1 & $c_z < 0$ & CG below foot radius \\
2 & $h_z > 0$ & hip above foot radius \\
3 & $h_x < 0$ & hip behind CG \\
4 & $\Delta_f > 0$ & foot radii displaced laterally \\
5 & --- & torque \& inertia break foot friction\\
\end{tabular}
\label{tab:rules}
\end{table}

\begin{table}[t]
\caption{Mugatu physical parameters}
\begin{center}
\begin{tabular}{ l | c c c }
Parameter & symbol & value & units \\
\hline
Mass & m & 809 & grams \\
Total height & $h$ & 18.5 & cm \\
Foot radius & $r$ & 12 & cm \\
CG Z offset & $c_z$ & -5.4 & cm \\
Hip Z offset & $h_z$ & 3.3 & cm \\
Hip X offset & $h_x$ & -1.4 & cm \\
Hip height & $r+h_z$ & 15.3 & cm \\
Foot gap & $\Delta_f$ & 3.2 & cm
\label{table:parameters}
\end{tabular}
\end{center}
\end{table}

\begin{figure*}[t]
\vspace{.25em}
\centering
\includegraphics[width=0.9\linewidth]{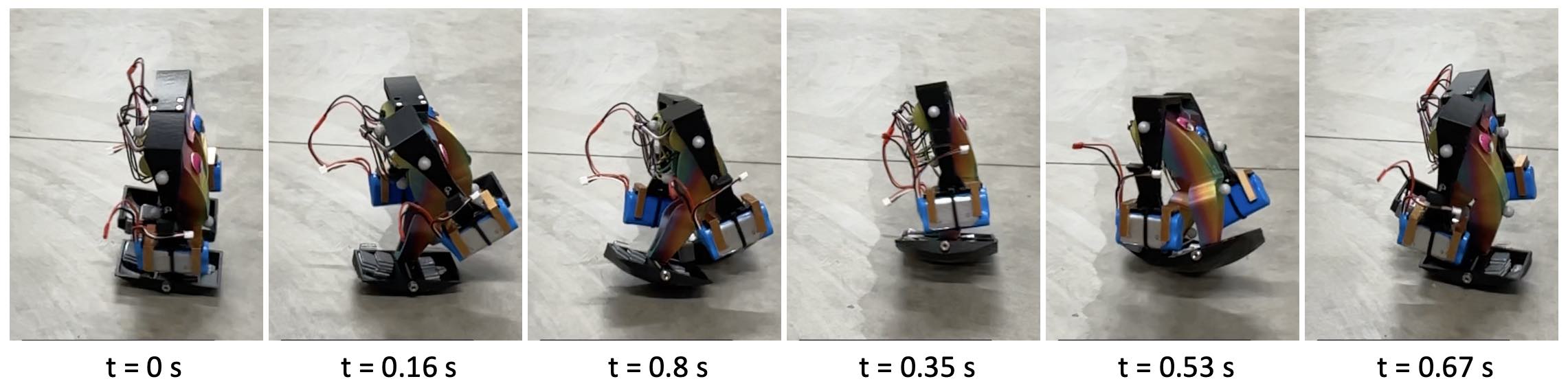}
\caption{Progression through a single step at a leg swing frequency of 1.4 Hz. Mugatu starts from legs together at 0s, raises it's left leg to begin walking, contacts the ground and rolls onto the left foot, swings the right leg through, and contacts the ground with the right foot completing a single gait cycle}
\label{fig:walkingDemo}
\end{figure*}

\subsection{How does it walk?}
\label{subsec:designRules}

To walk, a robot must lift and advance its feet while balancing and steering.
Usually, multiple motors coordinate these multiple tasks.
We show that careful mechanical design subject to several design rules enables one motor to conduct all of these tasks at once.
Fig.~\ref{fig:geometry} illustrates the design of the robot and the key parameters, with design rules given in Table~\ref{tab:rules} and the experimental robot's final parameters given in Table~\ref{table:parameters}.
This subsection describes how the robot lifts and advances its feet while balancing, and the following subsection describes how the robot can also steer.

We base our robot on classic passive dynamic walking toys like the Wilson Walker \cite{wilson1938walking}.
Feet with a large radius balance the robot by rolling it towards the upright position when tipped \cite{mcgeer1989wobbling}.
This passive stability requires satisfaction of Design Rule 1: the robot's center of gravity (CG) must be below the foot center of curvature ($c_z < 0$).
The feet must also be long enough to accommodate the range of tipping angles the robot excites during its walk so that it does not pivot over its ``heel" or ``toe" and fall down.

With such wide and relatively flat feet, the legs must be made long enough to lift them, leading to Design Rule 2: the hip joint must be above the foot center of curvature ($h_z > 0$).
When Rule 2 is disregarded, rotation of the hip joint will not effectively advance the swing leg.  The advancing foot may impact the ground as it advances, may tend to step in place, or may not be able to lift off of the ground at all depending on the exact location of the hip below the foot center of curvature.

In order for the robot to walk forwards, Design Rule 3 states that the hip axis must be displaced behind the CG and foot center of curvature when the robot stands upright with feet together ($h_x < 0$).
By displacing the hip axis backwards, we introduce an asymmetry that biases the robot to step forwards: when the hip is rotated away from zero degrees with legs aligned, the retreating leg is slightly lowered while the advancing leg is slightly raised, biasing the advancing leg to lift off the ground instead of the retreating leg.

Design Rule 4 is drawn from \cite{islam2022}: we identified that a positive gap between the hemispherical feet centers of curvature ($\Delta_f > 0$) leads to steady walking at a large range of frequencies -- a desirable quality in this work where the roll and leg swing motions are coupled and must both be excited by one motor. Note that the surfaces of the two feet are tangent at the bottom, and that they are not cut out of a sphere with the same center. 

Finally, Design Rule 5 states that the moment of inertia about the vertical axis and torque of the single motor must be high enough to break the friction of the feet standing on the ground in order to begin walking.
If these parameters are too low or the foot gap or friction are too large then the robot will spin in place without lifting its feet and will fail to walk. The robot would still be able to walk if this parameter were not satisfied, however, it would require a manual start and might yaw excessively with each step.

The general process of walking, shown in Fig.~\ref{fig:walkingDemo}, follows this sequence (assuming the robot takes a step first with its left leg for simplicity):
\begin{enumerate}
    \item The robot begins standing with feet together and its hip motor is commanded to follow a time-based oscillatory trajectory.
    \item Hip rotation lifts the left leg ahead of the robot due to Design Rule 3.
    \item With the left foot in the air and ahead of the right, the robot pitches forward and rolls left.
    \item The hip angle reaches its peak and the left leg begins to swing back, then contacts the ground.
    \item Due to the left leg's impact with the ground, the right leg lifts up.
    \item Since the robot has rolled left, the right leg has clearance to swing past the left, beginning the process again from step 3 with the roles of the legs swapped.
\end{enumerate}

\subsection{How does it turn?}
For a symmetric robot executing a symmetric gait, the robot can be expected to walk in a straight line.
To turn, the motor should introduce an asymmetry between the left and right steps.
Several asymmetries are possible: a faster leg swing velocity (or otherwise modified leg swing velocity profile) or a larger leg swing amplitude are two simple possibilities.
Regardless of method, the final result should be a longer step with one foot and a shorter step with the other.

\begin{figure}[t]
\centering
\includegraphics[width=0.9\linewidth]{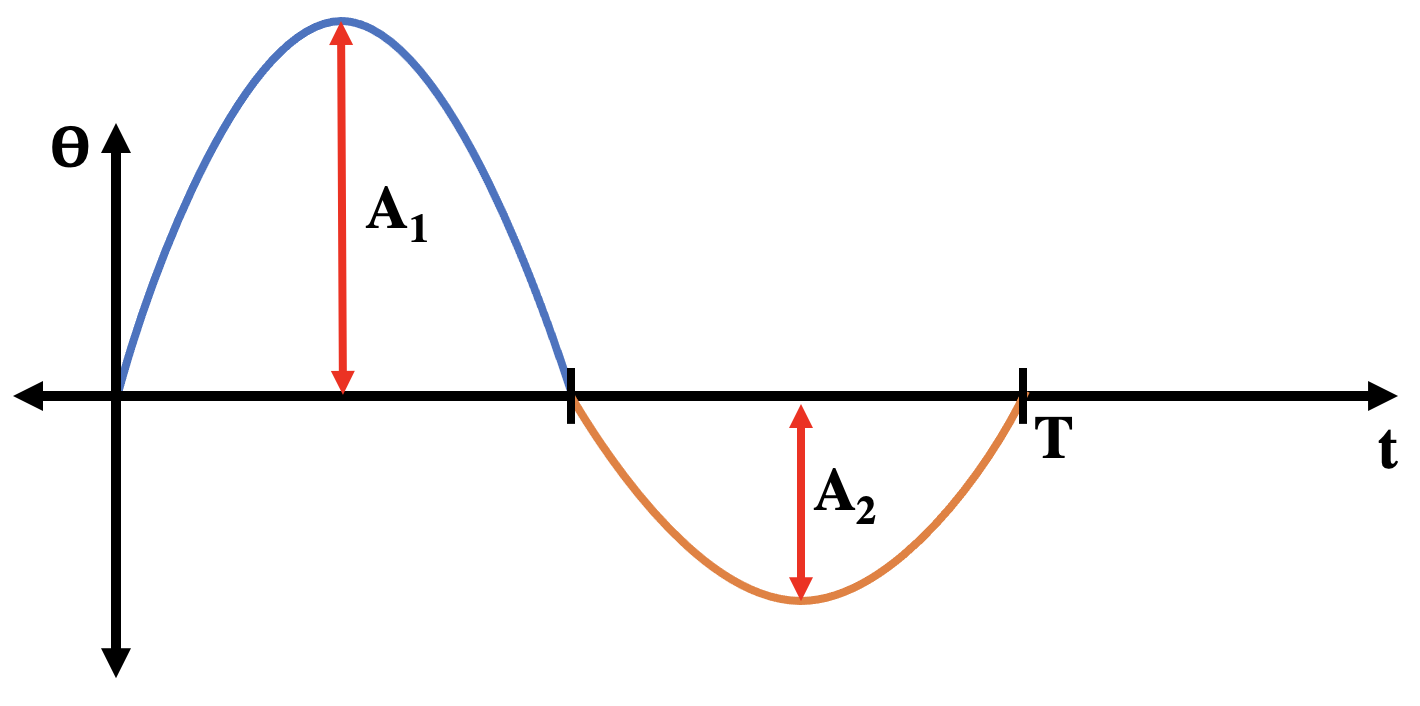}
\caption{Visualization of input signal with parameters labeled.}
\label{fig:inputSignal}
\end{figure}

Here, we control the legs using a piecewise sinusoid according to (\ref{eqation:InputSingal}) and visualized in Fig.~\ref{fig:inputSignal}. The sinusoid is broken at the midpoint into two sections, one half period for each leg swing. Each half has its own amplitude, \begin{math} {A_n} \end{math} but the same period. We demonstrate turning in Section~\ref{sec:results} by varying leg swing amplitudes asymmetrically with a larger peak on one side than the other.

\begin{align}
\theta (t) = \begin{cases}
  A_1 \sin(\omega t) & \text{mod}(t,T) < \frac{T}{2}\\
  A_2 \sin(\omega t) & \text{mod}(t,T) \geq \frac{T}{2}
\end{cases}
\label{eqation:InputSingal}
\end{align}

Interestingly, changing the breaking point between the slow and fast phases at the peak of the motion instead of the zero (that is, swinging one leg forward faster than the other by breaking the sinusoid at the extrema rather than the midpoints) did not result in consistent turning in hardware experiments.
The authors hypothesize that acceleration of the swing leg may play an important role in exciting yaw motion and that breaking the trajectory at the peaks produces phases of higher acceleration that split evenly across both swing phases rather than occur mainly on either the left or right side.

\subsection{Robot hardware}

The robot, ``Mugatu", Fig.~\ref{fig:robot}, consists of two rigid bodies connected by a single actuated hip joint. Each rigid body consists of one foot and leg rigidly connected to the opposing arm. The hip joint contains two bearings offset from each other to reduce friction and vibrations upon foot contact.

Each of the arms holds one battery for balance and to increase the moment of inertia about the vertical axis and move the leg CGs closer to the center line. While the arms are not included in the saggital model, we found that they were important to break the advancing leg's friction with the ground and lift it straight forward upon start. Furthermore, bringing the CG of each rigid body closer to the center reduces the coupled yaw motion when the legs swing. We used a Dynamixel XL330-M288-T servo to actuate the robot. The servo is located in the left leg with the right leg connected directly to the servo horn. We balanced this weight placement in the left leg by mounting the microcontroller and other electronic components in the right leg.

The feet are hemispherical in shape and the bottoms are coated in Plasti Dip to increase friction between the foot and the walking surface. The feet themselves are mounted with a lateral telescopic screw. With this, we can adjust the foot gap relatively easily and thereby adjust the natural roll frequency.
The feet hold wheel weights to satisfy Design Rule 1 keeping the CG low while maintaining a sufficiently small foot radius of curvature to satisfy Design Rule 2 keeping the hip above the foot radius.
These weights can be added or removed to tune the robot's walking.

Mugatu is powered by a single 7.4V LiPo battery (a second battery in the opposite arm is used as a weight and is not connected, though it could be used to extend the runtime) and controlled by an Arduino MKRZero microcontroller which commands the motors, collects data from a current sensor, and records experiment data to an SD card. A current sensor is connected in series with the battery to record Mugatu's overall current draw. There is also an MPU6050 IMU in the left leg for future feedback control, but it is unused in the testing for this paper.

Although the robot was designed according to the design rules mentioned in Section~\ref{subsec:designRules}, tuning is still required to produce stable walking. We broke the tuning process up into steps in order to tune out asymmetries in center of mass placement and find stable walking parameters. We started with a fixed foot gap, CG placement, and input amplitude. We then increased input frequency until the robot could settle into a stable walking gait. Once the robot could walk, we focused on tuning the walker for straight walking by changing center of mass placement in the y-direction. From experimentation, we found that biasing center of mass placement to one side caused the robot to turn preferentially towards that direction.

\subsection{Experimental setup}

We used two main systems for data collection on Mugatu. Motion capture records the position and orientation of both of Mugatu's rigid bodies. For every trial, we collected commanded leg swing angle, achieved leg swing angle, and current draw from the entire system saving the data to the SD card on board. Each trial was also video recorded with a 0.5x lens on an iPad. 

\begin{figure}[t]
\vspace{.5em}
\centering
\includegraphics[width=0.9\linewidth]{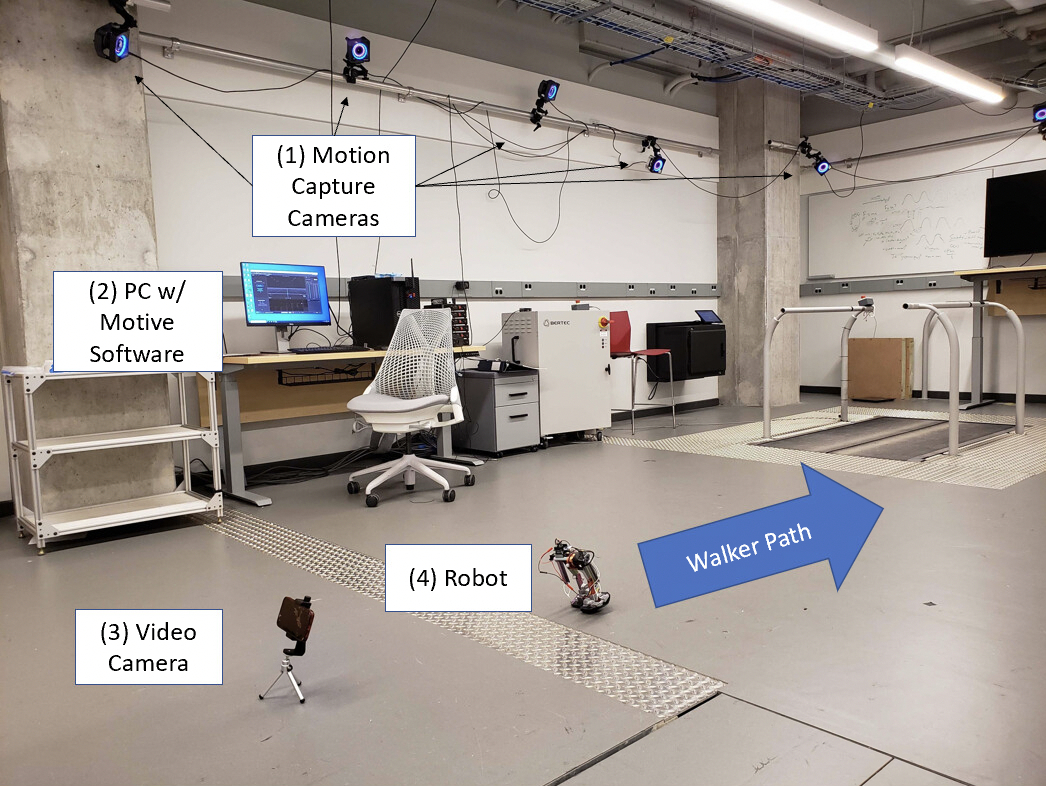}
\caption{Summary of the experimental environment. (1)  Optitrack motion capture cameras. (2) Optitrack Motive 3.0 motion capture software. (3) Video camera. (4) Robot. 
}
\label{fig:experimentalSetup}
\end{figure}

Fig.~\ref{fig:experimentalSetup} shows the layout of our experimental setup. Retro-reflective spherical markers allow motion capture to record the position and rotation of each leg-and-arm rigid body of the robot for later processing. The experimental environment is a smooth hard surface. The walker path shown above was used for all experiments for consistency and to get accurate camera tracking.

\section{Results}
\label{sec:results}

\begin{figure}[t]
\vspace{.5em}
\centering
\includegraphics[width=1.0\linewidth]{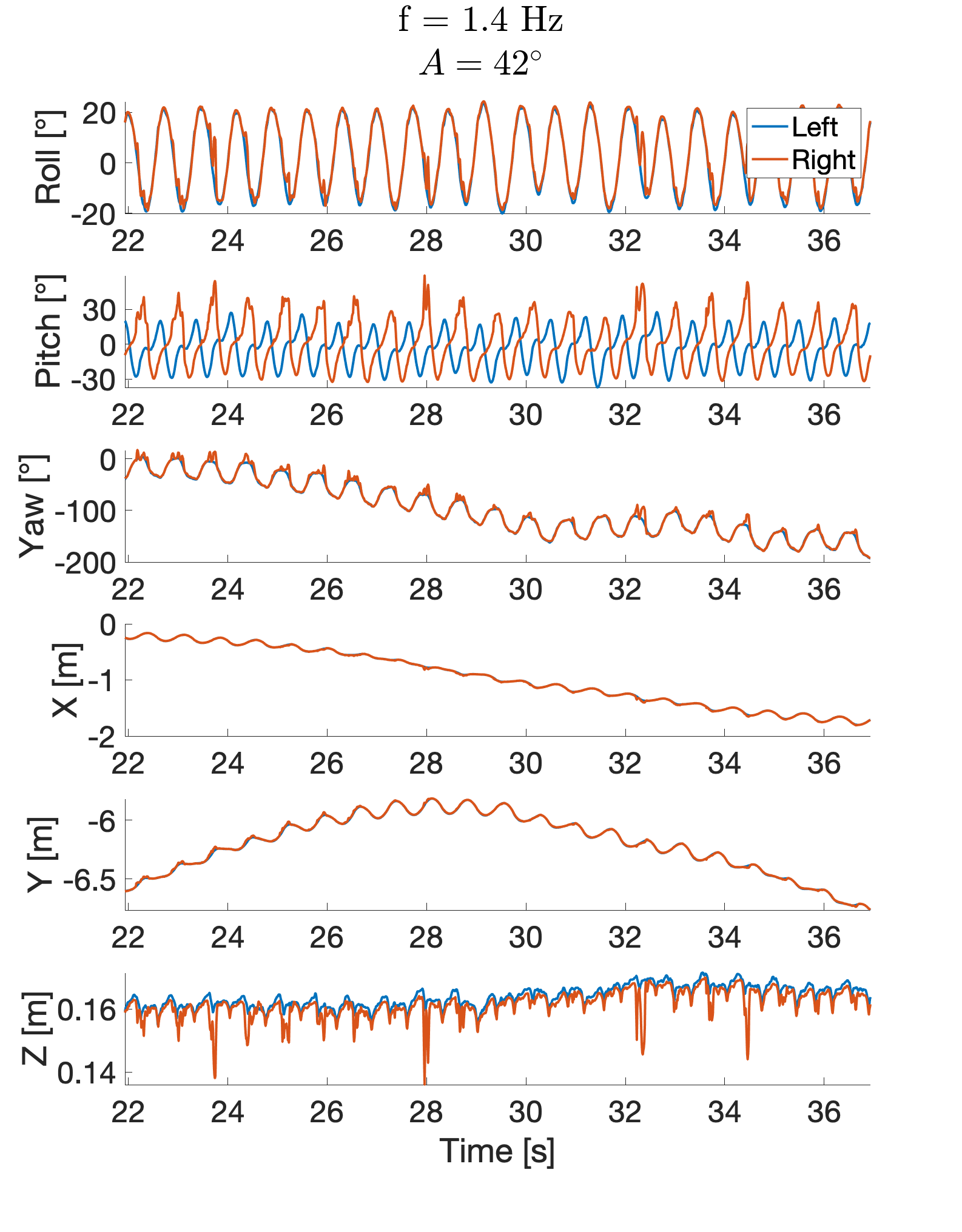}
\caption{Position and orientation data from an example trial with frequency 1.4 Hz and amplitude 42$^\circ$.}
\label{fig:exampleDataSet}
\end{figure}

To characterize the performance of this robot, we focused on its forward walking speed, turning capabilities, and efficiency. We also looked at the effect of parameter variations on these properties.

For each trial, we defined 3 time regions. The first region between the start time and first movement is the initialization process for the robot. Once first leg swing occurs, we assume a 10 second grace period, which we define as the second region, to allow the robot to reach steady state. Finally, we assume the robot walks steadily until the trial is ended or the robot is disturbed in some manner. We use this last region as our region of interest for analysis. Fig.~\ref{fig:exampleDataSet} illustrates example data from this region for a single trial. For most trials, we used the first 15 seconds within this time frame (that is, from 10 seconds to 25 seconds after first movement). For a few trials, the robot walked out of view of the motion capture system within that time period and we limited the data to when the robot was in view.

\subsection{Speed}

\begin{figure}[t]
\centering
\includegraphics[width=1.0\linewidth]{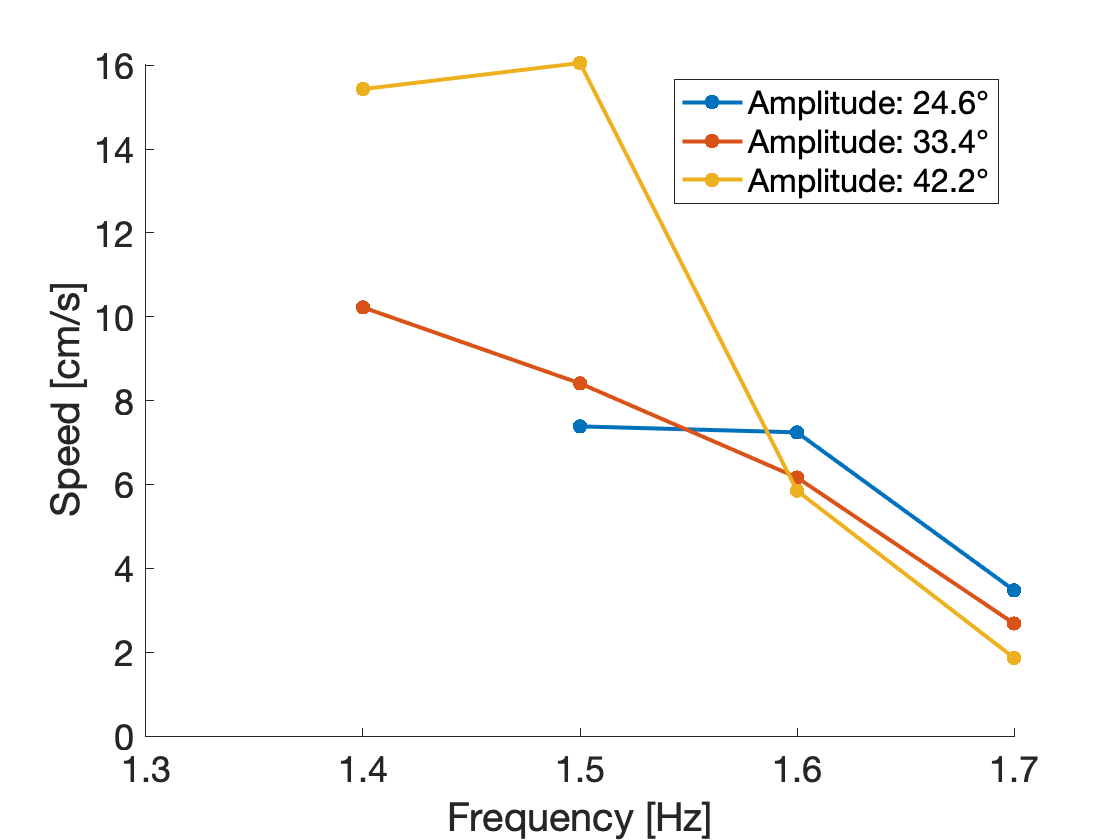}
\caption{Robot walking speed as a function of input frequency at varying leg swing amplitudes.} 
\label{fig:FrqVsAmpVsSpd}
\end{figure}

\begin{figure}[t]
\centering
\includegraphics[width=1.0\linewidth]{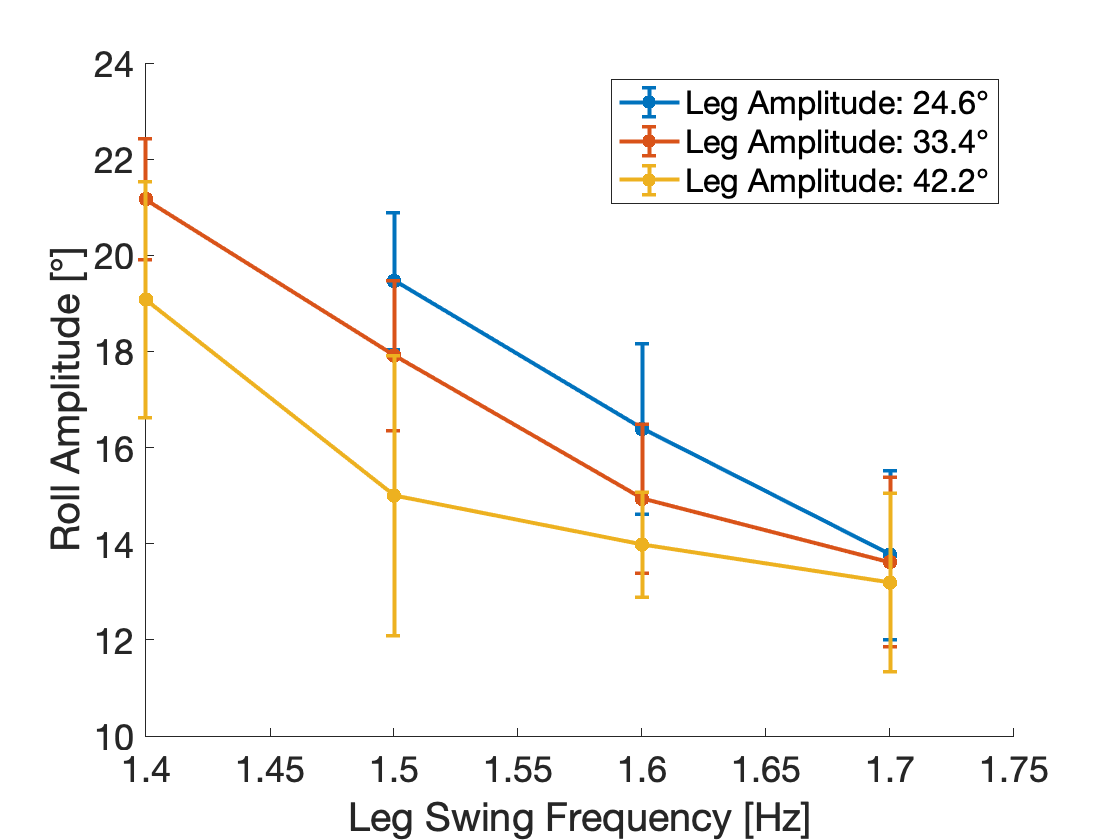}
\caption{Roll amplitude vs. leg swing frequency for varying leg swing amplitude. Error bars represent the standard deviation of peaks for each set of roll data used for analysis.} 
\label{fig:rollAmpVsLegSwingFrq}
\end{figure}

\begin{figure*}[t]
\vspace{.5em}
\centering
\includegraphics[width=1.0\linewidth]{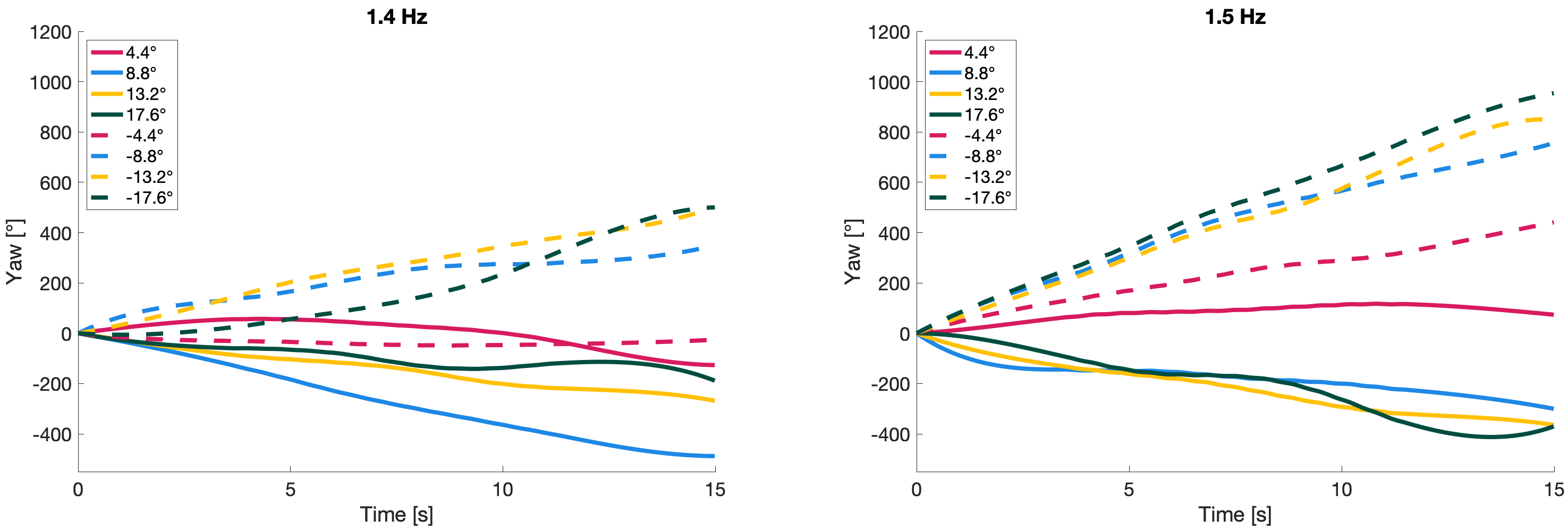}
\caption{Mid-line of the robot yaw for each amplitude difference and each frequency tested. Here, the legend is in terms of amplitude difference ($A_L - A_R$). Both legs start at a base amplitude of 33.4$^\circ$ and the absolute value of the amplitude difference in the legend is added to the left leg for the solid lines and the right leg for the dashed lines. Positive amplitude difference corresponds to a larger left leg swing.}
\label{fig:yawVsTime_Zoomed}
\end{figure*}

To characterize forward speed and walking stability, we examined the effects of symmetrically changing amplitude and frequency of the commanded sinusoidal trajectory, shown in Fig.~\ref{fig:FrqVsAmpVsSpd}. We chose a range of frequencies and amplitudes to experiment over to capture a range of walking behaviors both stable and unstable. The robot achieved a maximum walking speed of 16 cm/s at a commanded leg swing frequency of 1.5 Hz and a commanded leg swing amplitude of 42$^\circ$.

There are a couple of things to note from the results. First is the leg swing frequency threshold at which walking transitions from unstable to stable. We tested a range of frequencies from 1.3 Hz to 1.7 Hz for each amplitude line represented in Fig.~\ref{fig:FrqVsAmpVsSpd}. The first point for each amplitude line represents the first point at which stable walking occurred. The unstable trials (i.e. frequencies tested that do not have a point associated with them) mark the lower bound of leg swing frequencies at and below which roll oscillation and leg swing do not synchronize to produce stable walking. Second is the relationship between frequency and waking speed. We expect to see an optimal frequency for each amplitude of leg swing where the leg swings backwards less before touchdown and the stride length is larger. As shown in Fig.~\ref{fig:FrqVsAmpVsSpd}, the highest achieved speed for each amplitude is around the threshold of stability (i.e.\ around the lowest stable operating frequency). This is similar to the trends in \cite{islam2022}.

Another interesting relationship is that between roll amplitude and leg swing. We looked at the average roll amplitude from the region of interest for each trial (one trial per data point) in comparison to the leg swing frequency for that respective trial. The error bars in Fig. ~\ref{fig:rollAmpVsLegSwingFrq} refer to the standard deviation in peak height. Fig.~\ref{fig:rollAmpVsLegSwingFrq} shows that increasing leg swing frequency decreases roll amplitude. 
This is expected since a smaller period in leg swing translates to a smaller roll period which should translate into smaller roll amplitude for non-concentric hemispherical feet.
More interesting is the relationship between leg swing amplitude and roll amplitude. Intuitively, increasing leg swing amplitude might increase roll amplitude because a longer step might be expected to involve a larger roll. Fig.~\ref{fig:rollAmpVsLegSwingFrq} suggests the opposite trend. This could be a result of increased angular velocity of the swing leg translating into a higher inertial force pulling the robot to center.


\subsection{Turning}

We control the heading on Mugatu by introducing asymmetry into the amplitude of the input signal. The positive peak corresponds to the portion of the stride in which the left leg is in front of the right.  Making one hump larger than the other produces a difference in stride length between the two legs. Results from a sweep of amplitude differences in both directions at two different frequencies can be found in Fig. \ref{fig:yawVsTime_Zoomed}. The legs started at a base amplitude of 33.4$^\circ$ and the amplitude of one leg was increased to create a difference between the two legs. The amplitude difference shown in the legend was defined such that the left leg amplitude was subtracted from the right leg amplitude meaning a positive difference corresponds to a larger left leg swing. Because we are interested in turning rate and the raw yaw signal has an oscillatory aspect over small time scales, the lines in Fig. \ref{fig:yawVsTime_Zoomed} represent the midpoint of the periodic yaw signal for each step.
We were successfully able to control heading in both left and right directions, with more variability occurring at lower frequencies and more consistency at higher frequencies. At a frequency of 1.5 Hz, increasing the amplitude difference from 0$^\circ$ to 4.4$^\circ$ and 8.8$^\circ$ shows a clear trend of smoothly tightening turning radius. This relationship between yaw rate resolution and walking frequency could help inform future feedback controllers for path following. There also seems to be a bias towards left turning, especially at higher frequencies which could be due to mechanical asymmetries still present after tuning. We were also able to control heading to be relatively straight as demonstrated by the yaw rate at a frequency of 1.4Hz and frequency difference of -4.4°.

\subsection{Efficiency}

\begin{figure}[t]
\centering
\includegraphics[width=1.0\linewidth]{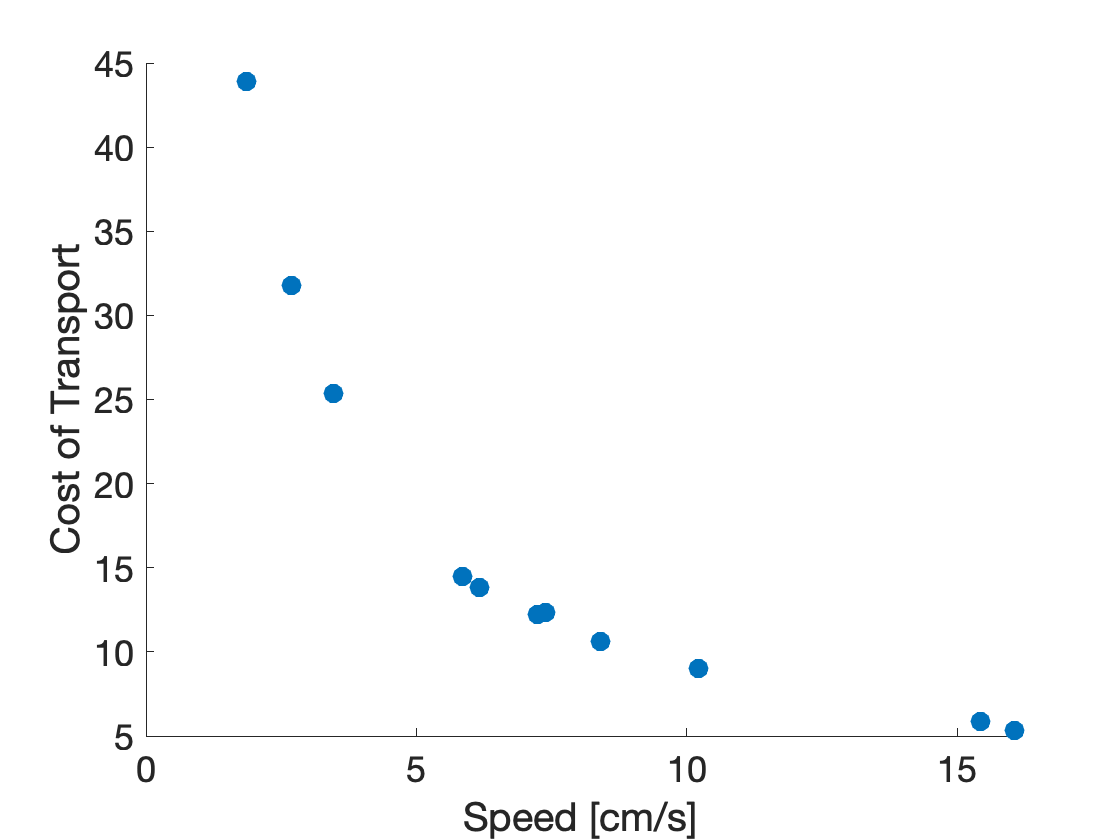}
\caption{Robot cost of transport as a function of walking speed.} 
\label{fig:TCoTVsSpeed}
\end{figure}

Cost of transport (CoT) is a nondimensional measure of efficiency defined as the energy expended to travel a certain distance divided by that distance times the system's weight \cite{tucker1975}. While Mugatu's electronics were not designed for efficiency, it is still an important value in investigating the feasibility of scaling down since size and weight constraints limit battery capacity at smaller scales. Using the data collected from the varying amplitude and frequency test in Fig.~\ref{fig:FrqVsAmpVsSpd}, we investigated the relationship between speed and resultant cost of transport. Based on the results in Fig.~\ref{fig:TCoTVsSpeed}, we can see that walking at higher speeds leads to higher efficiency and that the best CoT is 5.3. This is significantly higher than other robots at similar scales, e.g.\ iSprawl has a cost of transport around 1.75 \cite{iSprawl}. The variation in cost of transport with speed in Fig.~\ref{fig:TCoTVsSpeed} suggests that this inefficiency could be the result of non-motor related electronics dominating power consumption or that motor power requirements change little across the tested parameter ranges. The inverse relationship between speed and cost of transport indicates that CoT is dominated by the one over speed term and thus power consumption stays more or less constant with variation in both amplitude and frequency.


    



\addtolength{\textheight}{-3.5cm}   
                                  
\section{Conclusion}
The bipedal robot presented in this paper is possibly the simplest self-contained walking robot: 2 rigid bodies connected by one actuated revolute joint.
Despite walking involving foot lifting and advancing, side-to-side swaying, balancing, and steering, the single actuator successfully excites and controls coupled rolling, pitching, and swinging motions that produce an open-loop stable walking gait.
The robot demonstrates self-starting and stable walking at a range of speeds and walking frequencies.
Despite its morphological simplicity and single actuator, it is also able to steer left and right and smoothly vary its turning radius.
With the IMU already mounted on Mugatu, these turning capabilities could inform feedback controllers for straighter walking and path following in the future.

The robot's best cost of transport of 5.3 makes it less efficient than terrestrial animals of similar scale \cite{tucker1975}, but future design optimization studies could investigate if efficiency could be improved.
The robot's most efficient walking occurs at the highest speeds when the robot uses slow step frequencies with long strides.
However, the robot may have a larger stability margin at slightly higher leg swing frequencies since it falls over and fails to walk at frequencies below a threshold and experiments seem to show slightly more consistent turning at slightly faster frequencies.
This tradeoff between slow, efficient steps and faster more stable steps is an interesting relationship that was not investigated in depth during this study and could be a promising avenue for future work.

In this work, we focus on developing Mugatu as a new platform and exploring it's capabilities.  Future work with this platform could dive deeper into modeling of the robot's dynamics in both walking and turning.
This could lead to a better understanding of parameter relationships to important characteristics like walking speed, stability regions, and how those characteristics scale to inform future robot design.

The single-actuator hip-driven walking demonstrated here is more difficult to tune in comparison to leg extension actuation demonstrated before \cite{islam2022}.
However, the dramatic reduction in components and entire removal of sliding prismatic joints make it a promising morphology for designing even smaller walking devices.








\bibliographystyle{IEEEtran}
\bibliography{references.bib}

\end{document}